# Window-of-interest based Multi-objective Evolutionary Search for Satisficing Concepts

Eliran Farhi and Amiram Moshaiov

School of Mechanical Engineering

Tel-Aviv University

Tel-Aviv, Israel

*Abstract*— The set-based concept approach has been suggested as a means to simultaneously explore different design concepts, which are meaningful sub-sets of the entire set of solutions. Previous efforts concerning the suggested approach focused on either revealing the global front (s-Pareto front), of all the concepts, or on finding the concepts' fronts, within a relaxation zone. In contrast, here the aim is to reveal which of the concepts have at least one solution with a performance vector within a pre-defined window-of-interest (WOI). This paper provides the rational for this new concept-based exploration problem, and suggests a WOI-based rather than Pareto-based multi-objective evolutionary algorithm. The proposed algorithm, which simultaneously explores different concepts, is tested using a recently suggested concept-based benchmarking approach. The numerical study of this paper shows that the algorithm can cope with various numerical difficulties in a simultaneous way, which outperforms a sequential exploration approach.

## I. Introduction

The aim of this study is to devise and demonstrate a novel evolutionary search approach, which aims to find satisficing conceptual solutions. It concerns the Set-Based Concept (SBC) approach. In this approach, a conceptual solution (or in short – a concept) is a meaningful set of potential solution alternatives, which possess some common features [1, 2]. The considered concepts are predefined by the designers.

Fig. 1 illustrates the SBC approach. Three concepts of bridge designs are shown. The (generally different) design spaces of the concepts are marked by ellipses of different gray levels. As shown in Fig. 1, the associated performance vectors of particular designs, from all concepts, are to be compared in a mutual objective space. The most studied SBC approach is known as the s-Pareto approach [2]. It involves finding which particular designs, of which concepts, are associated with the global Pareto-front that is obtained by domination comparisons among all individual designs from all concepts.

It is conceivable that, using a meaningful division procedure, thousands of meaningful concepts may be constructed for a given real-life engineering problem. For example, we are currently working on the design of a propulsion system for unmanned aerial vehicles. For this particular problem, we have identified over 10,000 meaningful concepts that are of interest to the designers. It is noted that a non-interactive exploration run with 100 such concepts, using the algorithm of [3] may take about a week (on a standard workstation). In general, with an increasing number of concepts, finding the fronts of all the predefined concepts could be computationally prohibitive. This serves as a strong motivation for the work that is presented here.

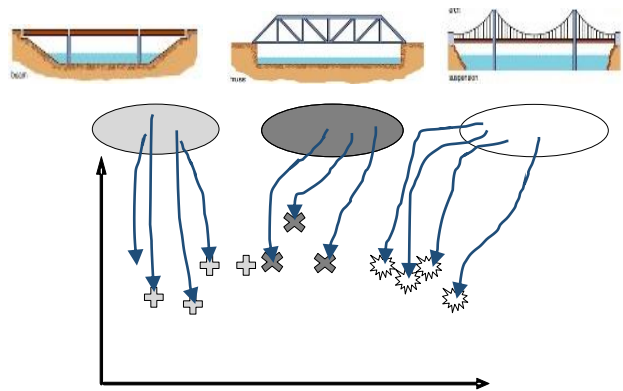

Figure 1: Illustration of the SBC approach

In [1], it was suggested that concepts should not be selected by the s-Pareto approach. A possible alternative has been suggested in [3], which employs a concept-based relaxed-Pareto approach.

In contrast to front-related methods, as in [1-4], the current work takes a Window-Of-Interest (WOI) approach. The WOI indicates what is considered as acceptable performance vector. Rather than being interested in finding concepts' fronts, here the designers are interested in finding which of the considered concepts have at least one solution with performance vector within a pre-defined WOI. Concepts that meet this requirement are considered satisficing. It should be noted that aspiration levels are commonly used to articulate objective preferences in multi-objective optimization. Yet, in contrast to studies as reviewed in [5], here aspiration levels, by way of the WOI, refer to satisficing concepts rather than to particular solutions.

The significance of the proposed approach, as compared with the concept-based relaxed-Pareto method of [3], is two-folds. First, in the method of [3], the sought information involves finding all the concepts' fronts within the relaxation zone. When the number of concepts within that zone is large, the computational effort might become prohibitive. Second, the sought information in the concept-based relaxed-Pareto method is defined by the relaxation zone, which is unknown in itself. This causes a computational difficulty in estimating the dynamic allocation of computational resources among the considered concepts. These two deficiencies are eliminated in the current approach. First, in the new problem definition there is no need to find any front. Second, the WOI, within which solutions are sought, is pre-defined.

As suggested above, thousands of meaningful concepts can be envisioned for a real life-problem. With this respect, it is noted that the proposed problem definition should be understood as a means to preprocess the concepts. Namely, it aims to reduce the considered set of concepts into those that have performances within the WOI. Eventually, depending on the final goal of the exploration, further search might be needed when aiming at selection of a concept or a particular design. Such a search could be done to either all the set of the obtained satisficing concepts from the preprocessing stage (the current problem solution), or to a subset of these concepts.

The contribution of this paper is threefold including: a. the proposed new problem definition and its rationalization, b. the suggested search algorithm, and c. the study to assess the proposed algorithm.

The rest of this paper is organized as follows. In section 0 the considered problem is defined. Next, section III provides the details on the proposed evolutionary search. Section IV describes the numerical experiments, their results and analysis. Finally, section V outlines the conclusions of this paper.

## II. PROBLEM DEFINITION

As mentioned above, the main goal of the problem is to reveal which of the concepts have at least one solution within the pre-defined WOI. This approach is different from the one that is described in [3], since it focuses on finding satisficing concepts rather than finding their Pareto fronts.

Let $n_o$ be the dimension of the objective-space $\mathbb{R}^{n_o}$. Also, let $S$ be the set of the considered concepts, and $n_c$ be the number of concepts ($n_c = |S|$). Let $X_m \subseteq \mathbb{R}^{n_m}$ be the design-space of the m-th concept, where $n_m$ is the dimension of $X_m$. Also, define $f_m: X_m \to \mathbb{R}^{n_o}$ as the concept's vector of objective-functions. Furthermore, let $s_m$ be any particular design of the m-th concept, and let $x_{s_m}$ represents the design vector of $s_m$. Also, let $y_{s_m} = f_m(x_{s_m})$ represents the performance vector of $s_m$, where $y_{s_m} \in \mathbb{R}^{n_o}$.

Without loss of generality, for a minimization problem let the WOI of the objective space be defined as the set:

(1) $\quad WOI = \{g \in \mathbb{R}^{n_o} | C_k(g) \leq 0 \ \forall \ k = 1, \ldots, n_0\}$

In other words, the decision-makers define $n_0$ inequality constraints in the objective space, where each $C_k$ is a hyperplane in the objective space. These should be defined in a consistent way with the optimization problem such that there is no vector in the WOI that is dominated by a vector outside the WOI.

A satisficing concept is defined as a concept that has at least one solution with performance vector, which is a member of the set WOI. The subset of the satisficing concepts, $C_{sat} \subseteq S$, is defined as:

(2) $\quad C_{sat} = \{C_j \in S | \exists y_{s_j} \in WOI\}$

Given a set $S$ of $\beta$ concepts, the considered problem is to find within given computational resources, up to $l$ concepts that are satisficing, where $l < \beta$.

## III. THE EVOLUTIONARY SEARCH

This section describes the search methodology. In subsection III.A the main procedure to search for the satisficing concepts is described. The other three subsections, III.B-D, contain the descriptions of the required sub-procedures, which are used by the main procedure.

### A. Main Procedure

The pseudo-code for the main procedure is described in Table 1.

TABLE 1: PSEUDO CODE FOR THE ENTIRE SEARCH

| 1 | *Initialization* | | |
|---|---|---|---|
| | 1.1 | *Define* the problem (see section II) | |
| | 1.2 | *Set* $GQ_0$, $GQ_i$, and the total # of generations | |
| | | *Set* counters J=0 and $I_j=GQ_0$ for each concept j | |
| | 1.3 | *Set* the GA parameters | |
| | 1.4 | *For each concept:* | |
| | | 1.4.1 | *Initialize* random population of solutions |
| | | 1.4.2 | *Perform* WOI-based Individual Evaluation (see sub-procedure in subsection III.B) |
| 2 | *For each concept j, while $I_j$ >0 and J<l, **Do**:* | | |
| | 2.1 | *Perform* Selection and Reproduction (see sub-procedure in subsection III.C) | |
| | 2.2 | *If the concept is satisficing then set J=J+1* | |
| | 2.3 | *Set $I_j=I_j$-1* | |
| 3 | *Perform* Resource Allocation (see sub-procedure in subsection III.D) | | |
| 4 | *While stopping conditions are not met go to step # 2* | | |

The main procedure starts with initializations, including the generation and initialization quotas. The i-th generation-quota ($GQ_i$) is a predefined number of generations, where i=1, …, $n_{cat}$, and $n_{cat} \leq n_c$. These quotas are needed for the resource allocation procedure (see subsection III.D). Also pre-defined is an initialization-quota ($GQ_0$), which is a predefined number of generations to be applied for each concept, before the resource allocation procedure is to be performed. Next, per each concept, an initial population is randomly generated. This is followed by the evaluation of the individuals, as detailed in subsection III.B. Then, an evolutionary process is carried out for each concept for a fixed number of generations ($I_j=GQ_0$).

Following this early evolutionary process, an advanced evolutionary process is taken place in which the resource allocation procedure becomes effective. The stopping conditions for the process are met when either the predefined number of satisficing concepts is obtained, or if the predefined total number of generations is reached.

### B. WOI-based individual evaluation

In this sub-procedure, individuals of a concept population are evaluated. The j-th concept's population $\left(P_j^g\right)$, is the set of all the evaluated designs, at the current generation # $g$, which are associated with concept j. First, the performance vector for each individual is obtained. Next, a distance, $d_k^j$, is calculated for each k-th individual of the j-th concept. This distance is the minimal Euclidean distance between the WOI and the performance vector of the individual. Next, each such individual is assigned with a WOI-based rank, denoted as $rank_k^j$. This ranking, which is based on a predefined number of distance ranges, $n_r$, is calculated as follows:

$$(3) \quad rank_k^j = \left\lceil \frac{d_k^j - \min_{k \in P_j^g} d_k^j}{\Delta} + 1 \right\rceil$$

Where Δ is defined as:

$$(4) \quad \Delta = \frac{\max_{k \in P_j^g} d_k^j - \min_{k \in P_j^g} d_k^j}{n_r}$$

After assigning a ranking level to each solution, as illustrated in Fig. 2, a crowding distance is calculated for individuals having the same rank.

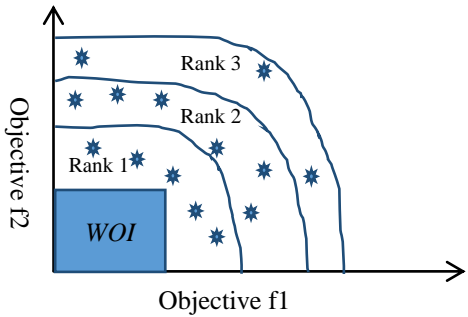

Figure 2: Illustration of ranking

The crowding-distance allows diversity preservation within the same rank. First, the boundary solutions are assigned by an infinite distance value, and then all other solutions are assigned with a calculated crowding distance. This procedure follows the one in [6]. The role of this assignment is to keep diversity and prevent premature convergence.

### C. WOI-based Selection and Reproduction

This sub-procedure, which is outlined in Table 2, is employed for each concept. It starts with tournament selection. This is based on lexicographic selection, where the rank is preferred over the crowding. If the ranks of the compared individuals are the same, then their crowding distance is employed. Next, the parent and the offspring populations of the considered concept are united, and the ranking and crowding distance are being calculated according to the sub-procedure in subsection III.B. Finally, a new elite population is selected using the ranks and the crowding distance.

TABLE 2: THE WOI-BASED SELECTION AND REPRODUCTION

| 1 | Tournament selection of the parent population |
|---|---|
| 2 | Create offspring population |
| 3 | Create the union population (parents & offspring) |
| 4 | Assign rank and crowding distance (see III.B) |
| 5 | Create the elite population |

### D. Resource Allocation

The purpose of this procedure is to distribute, during the search, the computational resources among all the concepts. The heuristic, which governs the resource allocation for each concept, is based on the expectation that the concept will be found to be satisficing. The resource allocation for each concept depends on its category, where those that belong to a better category (more promising) will get more resources than those with a worse category. The concept categorization, which is updated during the search, depends on the rate by which the concept solutions are approaching the WOI. Let $d^j(g)$ be the distance for the j-th concept, at generation # $g$, as follows:

$$(5) \quad d^j(g) = \min_{k \in P_j^g} d_k^j$$

The concept-distance-rate is the change of the concept-distance between generations, namely:

$$(6) \quad \Delta d^j = d^j(g) - d^j(g-1)$$

The concepts are sorted according to their concept-distance-rate and each is assigned with a category. Concepts, which are categorized into the 1st category (most-promising), receive a larger generation quota ($GQ_i$) as compared with those that belong to the lower categories. Resource allocation is recalled, for re-assigning the concepts with new generation-quotas, after all concepts finish their generation quotas.

## IV. EXPERIMENTS AND RESULTS

This section provides the results of testing the suggested algorithm and their analyses on min-min bi-objective problems using a WOI, which is defined by a minimal performance level for each objective. Subsection IV.A contains results from running the algorithm with only one concept at a time. Next, in subsection IV.B, results are provided concerning a benchmark example of running nine different concepts simultaneously. It follows the concept-based benchmarking method of [3]. Finally, in subsection IV.C, the same example is reworked with a different WOI.

### A. Single concept tests

This subsection describes some typical results from runs in which only one concept participate (one test function at each run). Namely, it focuses on the performance of the WOI-based selection, rather than on the performance of the resource allocation procedure. The latter is dealt with in the next two subsections.

To test the behavior of the algorithm, each of the nine bi-objective test functions, which are detailed in subsection IV.B, were employed with different population sizes and with different WOIs. The maximum number of generation was set to 1000, which is taken to be larger than needed for the particular examples.

To obtain statistical results, each case was tested for 30 runs. The runs were done by real-coding with crossover and mutations as in [6]. The crossover probability was taken as 0.9 and the mutation probability was taken as 1/n.

Table 3 provides typical results. The first column lists the employed test-function. The second and the third columns contain the WOI limits. The fourth column includes the population size. The fifth and sixth columns show the average number of generations, and total number of evaluations, until finding the concept to be satisfying.

TABLE 3: AVERAGE # OF GENERATIONS AND EVALUATIONS

| Test function | F1 limit | F2 limit | Pop. size | Avg. # of generations | Avg. # of Evaluations |
|---|---|---|---|---|---|
| ZDT1 | 0.5 | 0.5 | 100 | 244 | 24,400 |
|  |  |  | 50 | 319 | 15,950 |
|  |  |  | 20 | 436 | 8,720 |
|  |  |  | 10 | 556 | 5,560 |
| ZDT1 | 0.2 | 2.0 | 100 | 113 | 11,300 |
|  |  |  | 50 | 134 | 7,200 |
|  |  |  | 20 | 188 | 2,760 |
|  |  |  | 10 | 266 | 2,660 |
| ZDT1 | 2.0 | 0.2 | 100 | 193 | 19,300 |
|  |  |  | 50 | 210 | 10,500 |
|  |  |  | 20 | 272 | 5,440 |
|  |  |  | 10 | 334 | 3,340 |
| ZDT3 | 0.5 | 0.5 | 100 | 94 | 9,400 |
|  |  |  | 50 | 100 | 5,000 |
|  |  |  | 20 | 127 | 2,540 |
|  |  |  | 10 | 161 | 1,610 |

Based on the obtained results for all the considered concepts (test-functions), which are not shown here, it was decided to use a population size of 20 for each concept in the simultaneous searches of subsections IV.B-C.

Figs. 3-4 show typical population progress at different generations for runs with the ZDT1 test-function. The results are shown for two different WOI (gray color), using a population size of 20. In both cases, the same phenomenon is observed. At the beginning, the performances of the initial population are widely spread over the objective space. Then, the process exhibits a strong reduction of the diversity. Nevertheless, the search is not locked at local optima. This demonstrates the effectiveness of the suggested WOI-based selection approach. The fact that the search is successful, without a large diversity, calls for the use of a small population size or an adaptive population size.

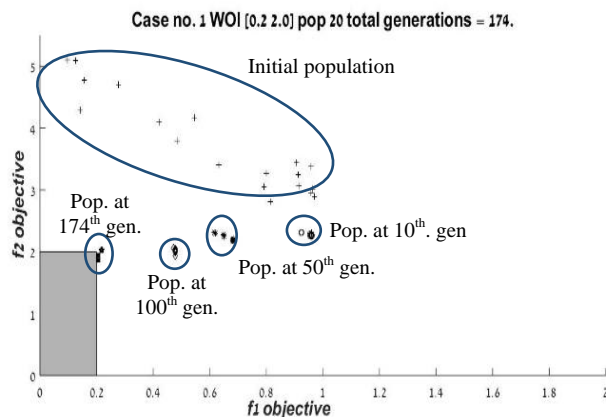

Figure 3: Search progress for ZDT1 with WOI = [0.2, 2.0]

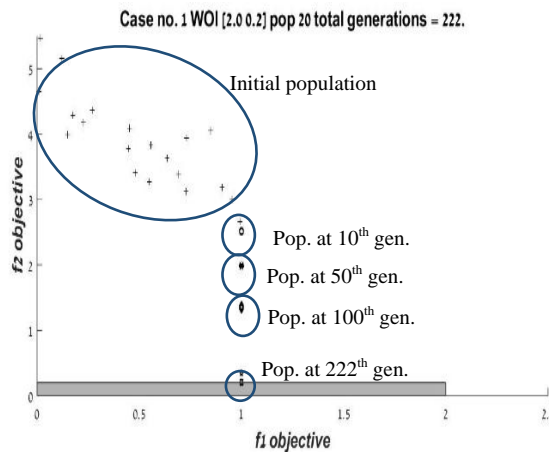

Figure 4: Search progress for ZDT1 with WOI = [2.0, 0.2]

Figure 5 describes the results for the case of ZDT1 with WOI = [0.5 ,0.5] and a population size of 20. It shows the statistics from 30 runs of the concept-distance vs. the number of function evaluations (in a boxplot form). In this particular case, the concept was found to be satisfying (distance reached a zero value) at around 8000 evaluations (for most runs).

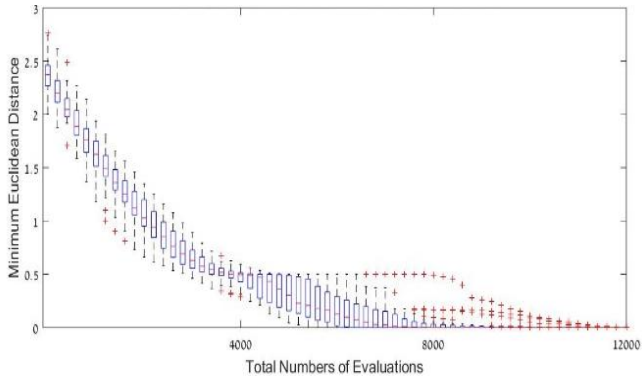

Figure 5: The statics of the concept-distance for ZDT1; WOI = [0.5 ,0.5]

*B. Simultanous search of concepts- Case 1*

This min-min bi-objective test problem involves nine concepts. Each concept is represented by a different test-function, based on well-know test-functions and their transformations (as in [3]). Table 4 shows all the concepts that were tested. The first column lists the function name and its version. For example: ZDT1-1 is for the classical ZDT1 problem and ZDT1-2 is a transformed ZDT1. The second and third columns correspond to the transformation scale and offset, respectively.

TABLE 4: THE DIFFERENT CONCEPTS DEFINITIONS

| Test Function | Objectives Scale | Objectives Offset |
|---|---|---|
| ZDT1-1 | [1,1] | [0,0] |
| ZDT1-2 | [2,2] | [-0.5,-0.5] |
| ZDT1-3 | [1,1] | [0.2,0] |
| SCH1-1 | [1,1] | [0,0] |
| SCH1-2 | [0.5,1] | [0,0] |
| ZDT2-1 | [1,1] | [0,0] |
| ZDT2-2 | [0.7,0.7] | [0,0] |
| ZDT2-3 | [0.7,0.7] | [0.2,0.2] |
| ZDT3-1 | [1,1] | [1,1] |

The Pareto fronts of the concepts are shown in Fig. 6. Also shown is the chosen WOI for this study case. It can easily be seen that only one concept (ZDT1-2) has solutions within the chosen WOI [0.2, 0.5].

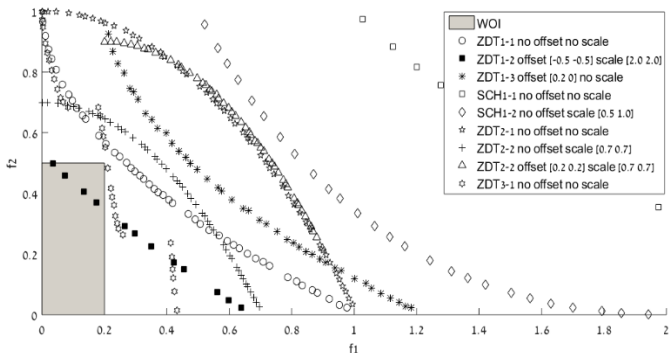

Figure 6: Pareto fronts of the concepts functions

In the resource allocation sub-procedure the concepts were categorized into three categories. The generation-quotas were set as GQ = [10,3,1]. In the current study, two of the concepts received the highest quota ($GQ_1=10$), five concepts received a medium quota ($GQ_2=3$), and the worst two concepts received the least quota ($GQ_3=1$).

In this case study the goal was to find just one satisficing concept, which is actually the only one such concept (ZDT1-2). Table 5 provides a comparison between sequential runs (10 runs for each concept at a time, with a limit of 1000 generations) and 10 simultaneous runs with the suggested resource allocation. The table shows the average number of generations that it took to stop the runs.

TABLE 5: PARALLEL VS. SEQUENTIAL RUNS FOR L=1, WOI = [0.2, 0.5]

| Concept # | Average # of generations for the simultaneous runs | Average # of generations in sequential runs |
|---|---|---|
| ZDT1-1 | 450 | 1000 |
| **ZDT1-2** | **498** | **499** |
| ZDT1-3 | 445 | 1000 |
| SCH1-1 | 142 | 1000 |
| SCH1-2 | 113 | 1000 |
| ZDT2-1 | 248 | 1000 |
| ZDT2-2 | 269 | 1000 |
| ZDT2-3 | 165 | 1000 |
| ZDT3-1 | 333 | 1000 |
| **Total** | **2663** | **8499** |

In all the simultaneous runs, the ZDT1-2 concept was revealed and the average amount of the total number of generations till stopping was 2,663 (which is equal to 53,260 evaluations). The sequential searches involved a total of 8,499 generations at most (= 169,980 evaluations). The latter number, and the comparison, should be understood as follows. In a random order of the sequential searches, the chance of choosing a satisficing concept is increasingly small with an increasing number of non-satisficing concepts. In such a case it appears that the simultaneous approach has a clear advantage over the sequential approach.

Fig 7 shows, for each concept, the median of the concept-distance versus the number of generations, as obtained by the simultaneous approach.

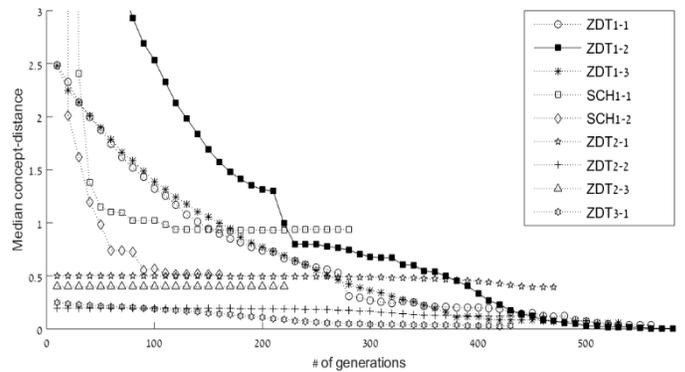

Figure 7: Medians of concept-distances versus # of generations

The statics of the number of generations till stopping, per each concept, are displayed in Fig. 8. As seen there, the less promising concepts received a relative small amount of resources during the search, whereas the more promising ones received a larger amount of resources, until the satisficing concept was found.

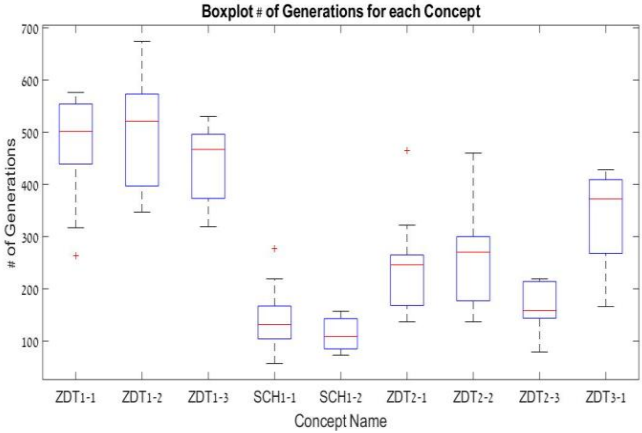

Figure 8: The statistics of the # of generations per concept

*C. Simultanous search of concepts- Case 2*

The second case study is almost the same as the first one. The only difference is that the WOI was re-set to [0.3, 0.4], which amounts to having two rather than one satisficing concepts (ZDT1-2 and ZDT3-1). The goal of this case was set to find the two satisficing concepts. The same comparison between a parallel run and a sequence run was made. The results are summarized in table 6.

The average number of the total generations used by the algorithm was 4,008 (which is equal to 80,160 evaluations), whereas in the sequential runs, the total number of generation was 8,036, which is equal to 160,720 evaluations). As expected, with increasing number of satisficing concepts, as compared with the previous case, the advantage of the simultaneous approach is reduced.

TABLE 6: PARALLEL VS. SEQUENTIAL RUNS FOR L=2, WOI=[0.3,0.4]

| Concept # | Average # of generations for the simultaneous runs | Average # of generations in sequential runs |
|---|---|---|
| 1 | 584 | 1000 |
| **2** | **530** | **605** |
| 3 | 620 | 1000 |
| 4 | 186 | 1000 |
| 5 | 167 | 1000 |
| 6 | 488 | 1000 |
| 7 | 488 | 1000 |
| 8 | 473 | 1000 |
| **9** | **472** | **431** |
| **Total** | **4008** | **8036** |

V. SUMMARY AND CONCLUSIONS

This paper describes a new type of a concept-based multi-objective problem, in which the decision-makers are interested in finding some satisficing concepts out of possibly a very large number of concepts. A satisficing concept is defined here as a concept with at least one particular solution within a window-of-interest in the objective space. An evolutionary algorithm is suggested for solving this problem by a simultaneous search with the considered concepts. The algorithm is examined using concept-based benchmarking problems, which simultaneously employ various well-known test functions. The simultaneous approach is compared with a sequential one. It is concluded from the current case studies that the algorithm performs well and is capable to reveal the sought satisficing concepts, while reducing the resources from non-promising concepts. The comparisons with the sequential approach suggest, as expected, that when the ratio of the number of satisficing concepts to that of the non-satisficing ones is decreased (first case study as compared with the second case), the benefit of using the simultaneous search approach is more evident.

A major element in making search algorithms efficient for the considered type of problems is to have a good on-line prediction technique that a concept is expected to be satisficing. With this respect, future studies should suggest and compare various such predictors. Future studies should also employ additional benchmark problems with different arrangements of the concepts fronts, more objectives and different WOIs. Developing and comparing of other evolutionary algorithms should also be investigated.

Finally, real-life engineering problems with increasing number of concepts may require the development of both parallel processing and interactive approaches. These are left for future research.